\documentclass[journal]{IEEEtran}
\usepackage{booktabs}
\usepackage{threeparttable}
\usepackage{cite}
\usepackage{indentfirst}
\usepackage{setspace}
\usepackage{CJK}
\usepackage{amsmath}
\usepackage{amssymb}
\usepackage{amsthm}
\usepackage{graphicx}
\usepackage{caption}
\usepackage{subcaption}
\usepackage{geometry}
\usepackage{float}
\usepackage{algorithm2e}
\usepackage{multirow}
\usepackage[symbol]{footmisc}

\begin{document}

\title{Interpretable End-to-end Urban Autonomous Driving with Latent Deep Reinforcement Learning}

\author{Jianyu Chen$^{1}$, Shengbo~Eben~Li$^{2}$, Masayoshi Tomizuka$^1$
\thanks{

$^1$Department of Mechanical
Engineering, University of California, Berkeley, CA 94720, USA.
\texttt{Email: jianyuchen@berkeley.edu, tomizuka@berkeley.edu}
\quad 

$^2$State Key Lab of Automotive Safety and Energy, School of Vehicle and Mobility, Tsinghua University, Beijing 100084, China. \texttt{Email: lishbo@tsinghua.edu.cn}}
}


\maketitle

\begin{abstract}

Unlike popular modularized framework, end-to-end autonomous driving seeks to solve the perception, decision and control problems in an integrated way, which can be more adapting to new scenarios and easier to generalize at scale. However, existing end-to-end approaches are often lack of interpretability, and can only deal with simple driving tasks like lane keeping. In this paper, we propose an interpretable deep reinforcement learning method for end-to-end autonomous driving, which is able to handle complex urban scenarios. A sequential latent environment model is introduced and learned jointly with the reinforcement learning process. With this latent model, a semantic birdeye mask can be generated, which is enforced to connect with a certain intermediate property in today's modularized framework for the purpose of explaining the behaviors of learned policy. The latent space also significantly reduces the sample complexity of reinforcement learning. Comparison tests with a simulated autonomous car in CARLA show that the performance of our method in urban scenarios with crowded surrounding vehicles dominates many baselines including DQN, DDPG, TD3 and SAC. Moreover, through masked outputs, the learned policy is able to provide a better explanation of how the car reasons about the driving environment. The codes and videos of this work are available at our github repo\footnote[2]{\texttt{https://github.com/cjy1992/interp-e2e-driving}} and project website\footnote[3]{\texttt{https://sites.google.com/berkeley.edu/interp-e2e/}}.
\end{abstract}

\begin{IEEEkeywords} 
Autonomous driving, Deep reinforcement learning, End-to-end driving policy, Probabilistic graphical model, Interpretability.
\end{IEEEkeywords}

\IEEEpeerreviewmaketitle

\section{Introduction}\label{sec:intro}

\IEEEPARstart{M}ost of today's autonomous driving systems are using a highly modularized hand-engineered approach, for example, perception, localization, behavior prediction, decision making and motion control, etc~\cite{thrun2006stanley,urmson2008autonomous}. Take the perception module as an example: even though some learning techniques are used, its design still needs tedious hand-engineered work like selecting representation features of each types of road users. Even though working well in a few driving tasks, this modularized framework starts to touch its performance limitation in urban driving scenarios because (1) too much human heuristics can lead to conservative driving policy; (2) it is hard to generalize as we might need to redesign the heuristics for each new scenario and task, and (3) these modules are strongly entangled with each other, and the whole system becomes expensive to scale and maintain. 

Those limitations might be avoided with end-to-end autonomous driving approaches, in which a driving policy can be learned and generalized to new tasks without much hand-engineered involvement~\cite{bojarski2016end,codevilla2018end,xu2017end}. Moreover, the learned policy can be continuously optimized in driving, which is possible to achieve superhuman performance. Two main branches for end-to-end autonomous driving are imitation learning (IL)~\cite{bansal2018chauffeurnet,bojarski2016end,chen2019deep,codevilla2018end}, which learns a driving policy by imitating the collected expert driving data, and reinforcement learning (RL)~\cite{wolf2017learning,lillicrap2015continuous,kendall2019learning}, which learns a policy by self exploration and reinforcement. However, existing end-to-end methods are criticized by two main shortcomings: 1) The learned policies are quite lack of interpretability because neural network is like a black-box. When a deep neural network is learned directly from raw observations to control command, we can not explain how it works. 2) They can only deal with simple driving tasks such as lane keeping or car-following. However, urban autonomous driving is much more complex due to highly dynamic road traffic and strong road user interaction. The various urban scenarios and street views significantly increase the sample complexity, making it extremely challenging to learn a good end-to-end driving policy.

This paper introduces the maximum entropy RL with sequential latent variables to address the problems in end-to-end autonomous driving. The latent space is employed to encode the complex urban driving environment, including visual inputs, spatial features, road conditions and road users' states. Historical high-dimensional raw observations are compressed into this low-dimensional latent space with a sequential latent environment model, which is learned jointly with maximum entropy reinforcement learning process. 

The introduced latent space enables an interpretable explanation of how the policy reasons about the environment by decoding the latent state to a semantic birdeye mask. During training, this mask is enforced to connect with some intermediate properties in today's modularized framework, for example, localization \& mapping, object detection, and behavior prediction, thus providing an explanation on the learned policy. Meanwhile, the latent space provides a much more compact state representation, which significantly reduces sample complexity of learning the driving policy, resulting in a large performance improvement. We implemented our method to learn an end-to-end driving policy from raw camera and lidar inputs in CARLA simulator. Experimental evaluation demonstrates that our method significantly outperforms prior methods in crowded urban scenarios. Examples of decoded semantic birdeye masks are presented to illustrate how our autonomous car understands the driving situations.

\section{Related Works}
Recent advances in machine learning enables the possibility of learning based end-to-end approaches for autonomous driving. There are two main approaches: imitation learning (IL) and reinforcement learning (RL). IL learns a driving policy from expert driving data~\cite{bojarski2016end,codevilla2018end,bansal2018chauffeurnet,chen2019deep}. With expert samples as labelled data, a driving policy is often easy to train, and it generally works well in structured driving tasks if one can collect enough expert data. However, there are fundamental limitations for IL: (1) IL is data hungry, and moreover its performance is limited to the average of the demonstration data; (2) IL is unable to learn skills that are not provided or rare in the demonstration data. This makes it difficult to deal with some dangerous scenarios such as near collision cases because they might never be demonstrated by the expert. 

Combined with deep learning techniques, RL shows its power on tackling complex
decision making and planning problems, bringing a series of breakthroughs in recent years. Agents trained with deep RL techniques achieves super-human-level performance in game playing~\cite{mnih2013playing,mnih2015human,vinyals2019alphastar}, go playing~\cite{silver2016mastering,silver2017mastering}, and robotics~\cite{levine2016end,kalashnikov2018qt}. Related deep RL algorithms range from value based methods such as DQN~\cite{mnih2013playing,mnih2015human} and double DQN~\cite{van2016deep}, actor-critic based methods such as A3C~\cite{mnih2016asynchronous}, DDPG~\cite{lillicrap2015continuous} and TD3~\cite{fujimoto2018addressing}, policy optimization based methods such as TRPO~\cite{schulman2015trust} and PPO~\cite{schulman2017proximal}, and maximum entropy RL methods such as SAC~\cite{haarnoja2018soft,haarnoja2018}. With RL, a policy can be learned automatically without any expert data. It can explore various kinds of possible cases including some dangerous ones, and then learn related skills. It also has the potential to achieve superhuman performance.

Researchers have been trying to apply deep RL techniques to the domain of autonomous driving. Wolf et al.~\cite{wolf2017learning} used DQN to learn to steer an autonomous car to keep in the track in simulation. Its action space is discrete and only allows coarse steering angles. Lillicrap et al.~\cite{lillicrap2015continuous} proposed a continuous control deep RL algorithm which learns a deep neural network policy that is able to drive the autonomous car on a simulated racing track. Chen et al.~\cite{chen2018deep} proposed a hierarchical deep RL framework to solve driving scenarios with complex decision making such as traffic light passing. Kendall et al.~\cite{kendall2019learning} demonstrated the first application of deep RL to real world autonomous driving. They learned a deep lane keeping policy using a single front-view camera image as input. There are a lot of other related works not mentioned here. However, existing works are either for simple scenarios without complex road conditions and multi-agent interactions, or use manually designed feature representations.

Another problem of learning-based approaches for autonomous driving is that they are lack of interpretability. The learned deep neural network policy is like a black box, which is not ideal since autonomous driving is a safety critical real world application. It is important for us to know whether and how the autonomous car understand the environment. Some works have made efforts in this direction. Bojarski et al.~\cite{bojarski2017explaining} visualized NVIDIA's deep neural network based driving system by extracting the convolutional layer feature maps and finding the salient objects. Kim et al.~\cite{kim2017interpretable} used a visual attention model with a causal filter to visualize the attention heatmap. Sauer et al.~\cite{sauer2018conditional} analyzed the decision making process of the deep neural network by using gradient-weighted class activation maps to obtain the attention of the CNN. However, the interpretable information they provide --- mostly just tell which part of the observed image is within attention --- is rather weak.

Probabilistic graphical model (PGM) is a generic and powerful tool to formulate many machine learning problems~\cite{murphy2012machine}. Sequential latent model~\cite{krishnan2015deep,karl2016deep,fraccaro2017disentangled,hafner2018learning,lee2019stochastic} is one of its very relevant applications to this work, which uses PGM to formulate stochastic time sequence processes with latent variables. Close connections are also found between PGM and maximum entropy reinforcement learning~\cite{levine2018reinforcement,rawlik2013stochastic,ziebart2010modeling}. Some recent works propose to integrate sequential latent model learning and reinforcement learning~\cite{lee2019stochastic,ha2018world,hafner2018learning,okada2020planet}. Such methods show great potential in end-to-end learning of deep policies with high dimensional inputs. However, no prior works have used this branch of techniques to formulate and solve autonomous driving problems. Furthermore, they do not provide interpretability of the learned model, and do not take muiltiple sources of sensor inputs, which is essential for autonomous driving systems.

\section{PGM for Environment Modeling and Reinforcement Learning}\label{sec:PGM}
\subsection{Probabilistic Graphical Model (PGM)}
\begin{figure}
  \centering
  \includegraphics[width = .175\textwidth]{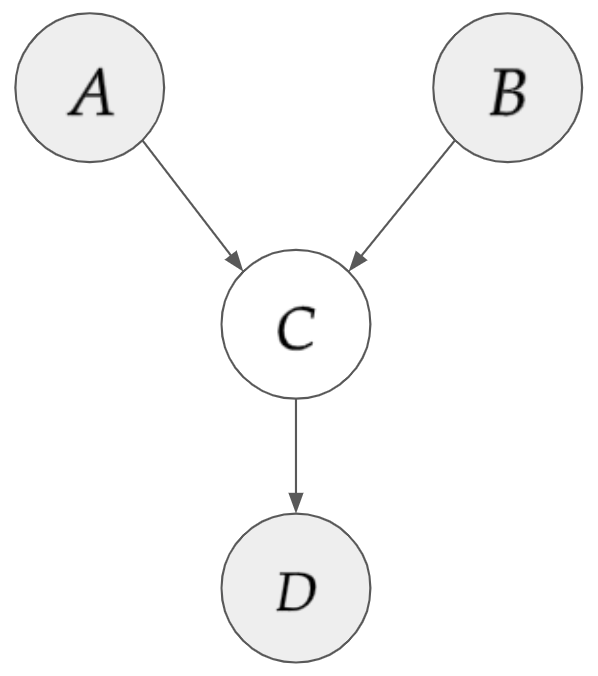}
  \caption{\label{Fig:PGM}A simple example of probabilistic graphical model}
\end{figure}

Probabilistic graphical model (PGM) is probabilistic, but uses a graph to represent conditional dependence between random variables~\cite{murphy2012machine}. They are widely used in Bayesian statistics and Bayesian learning. Fig.\ref{Fig:PGM} shows a simple example of PGM. There are in total 4 nodes A, B, C and D. These nodes can represent random variables meaning observable quantities, unobservable latents, or unknown parameters. The edges between nodes represents conditional dependencies. In Fig.\ref{Fig:PGM}, C is conditioned on A and B, while D is conditioned on C. Each edge is associated with a conditional probability, such as $p\left( {\left. C \right|A,B} \right),\;p\left( {\left. D \right|C} \right)$. With the ability to describe complex causal effects and probabilistic transitions, PGM can be used as a universal model in stochastic RL. We will now introduce how PGM is used to model a discrete-time dynamic environment.

\subsection{PGM for Sequential Latent Environment Modeling}
To obtain optimal policy, it is crucial to accurately model the environment. The environment in its nature has the following characteristics: (1) High dimensional observations: either for a human being or an autonomous car, the raw observations for them are usually high dimensional, such as RGB images; (2) Time-sequence probabilistic dynamics: the state of the environment will change with time, thus time sequence relations should be modeled; (3) Partially observable: the observation at the current time alone might not be enough to recover full state of the environment, historical information needs to be summarized by historical observations.

\begin{figure}
  \centering
  \includegraphics[width = .38\textwidth]{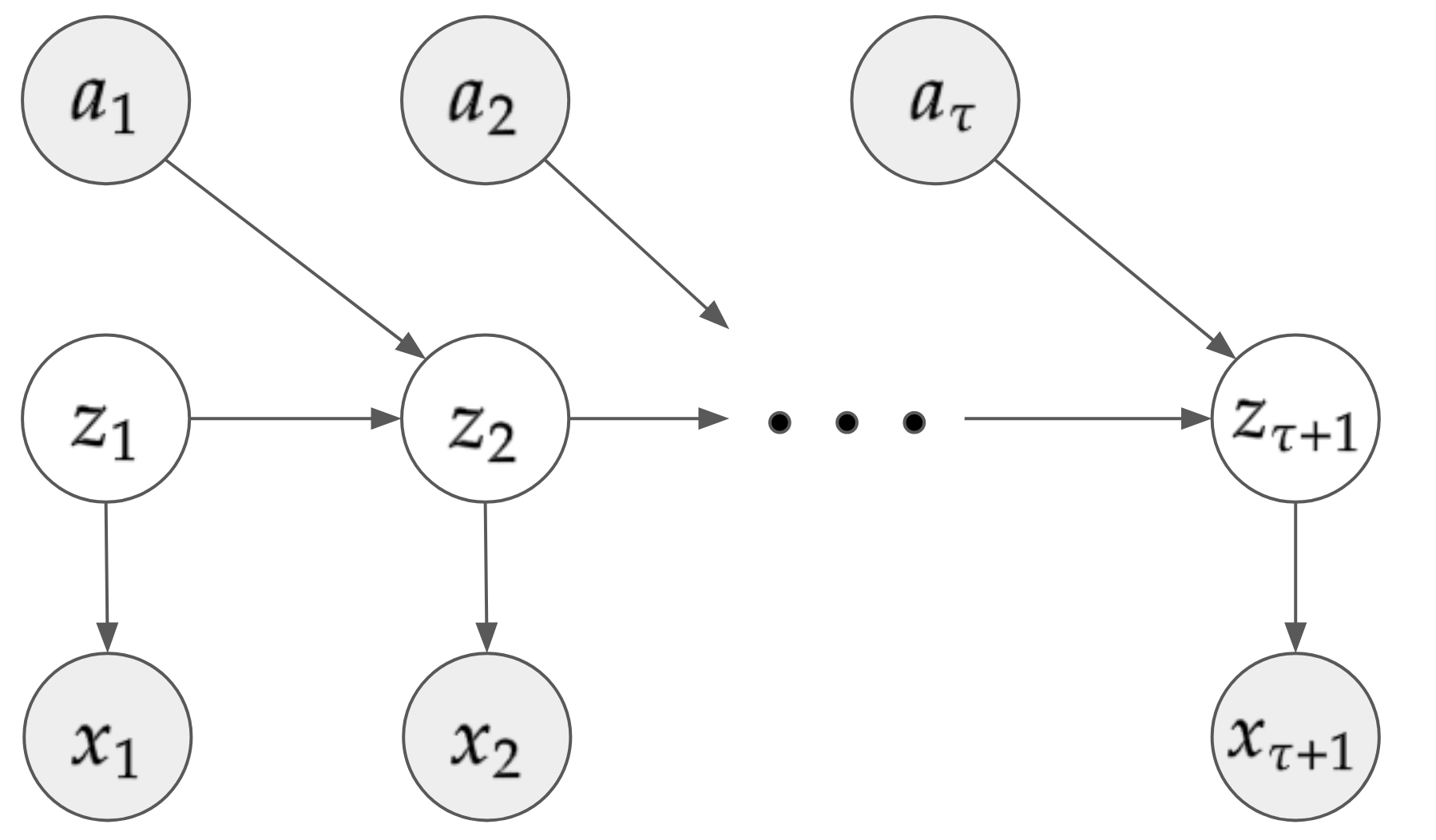}
  \caption{\label{Fig:PGMenv}A PGM sequential latent environment modeling}
\end{figure}

Here we introduce a probabilistic sequential latent environment model, which satisfies the above stated characteristics. Similar structures of this model is adopted by multiple literatures~\cite{krishnan2015deep,hafner2018learning,lee2019stochastic}. As shown in Fig.\ref{Fig:PGMenv}, $x_t$ represents the observation at time step $t$, which can be high dimensional sensor inputs such as RGB images. $a_t$ is the action chosen at $t$. $z_t$ is the latent state variable at $t$, which is a description of the current situation summarizing historical information, e.g, the position, velocity, intention of other road participants, the drivable areas, and the road markings. The observation $x_t$ is a decoding of the latent state $z_t$, defined by $p\left ( x_t|z_t \right )$. The latent state $z_t$ together with the action $a_t$, decide the latent state at the next time step by the state transition function $p\left ( z_{t+1}|z_t, a_t \right )$.

This environment model is quite generic, as there is no restrictions of the format and physical meaning of observation, action, and latent state. Furthermore, the observation decoding function $p\left ( x_t|z_t \right )$ and state transition function $p\left ( z_{t+1}|z_t, a_t \right )$ can be arbitrarily complex, such as deep neural networks.

By introducing an additional filtering function $p\left ( z_{t+1}|z_t,x_{t+1},a_t \right )$, the latent state can be inferred in a recursive bayesian filter way. Given a new observation $x_{t+1}$, we have $p\left(z_{t+1} \right )= p\left ( z_{t+1}|z_t,x_{t+1},a_t \right )p\left (z_{t} \right )$, where $a_t$ is the action executed at the last time step. The latent state for the first time step is obtained by $p\left(z_1 \right )=p\left(z_1|x_1 \right )$. We can also make probabilistic predictions by rolling out the future states based on the state transition function:
\begin{equation}
    p\left(z_{\tau:\tau+H}|a_{\tau:\tau+H-1} \right )=p\left(z_\tau \right )\prod_{t=\tau}^{\tau+H-1} p\left(z_{t+1}|z_t,a_t \right )
\end{equation}

Furthermore, with the decoding networks, we can not only decode to the raw observations for unsupervised learning, but can also decode to any other representations, such as a semantic mask to provide interpretable explanations.

We can fit the parameters $\psi$ of this PGM from dataset, which is composed of observation-action trajectory sequences ${\cal D} = \left \{ \left(x_{1:\tau}^i, a_{1:\tau}^i \right ) \right \}_{i=1}^N$, by maximizing the likelihood of the data:
\begin{equation}
    \underset{\psi}{\mathrm{max}}\; \prod_{i=1}^N p\left (x_{1:\tau}^i|a_{1:\tau}^i \right ) 
\end{equation}

\subsection{PGM for Reinforcement Learning}
Under the settings of reinforcement learning~\cite{sutton2018reinforcement}, at each time step, an agent observes the state $z_t$, executes action $a_t$ generated by its policy $a_t \sim \pi\left(a_t|z_t \right )$, and then gets the reward $r\left(z_t,a_t \right )$. The state is then updated according to the state transition $z_{t+1}\sim p\left ( z_{t+1}|z_t, a_t \right )$. Assume there are $H$ time steps in an episode and the initial state is generated by $z_1 \sim p\left(z_1 \right)$, then the objective of reinforcement learning is to find an policy that optimizes the expected accumulative rewards:
\begin{equation}
    \pi^* =\underset{\pi}{\textrm{argmax}}\underset{\substack{z_1 \sim p(z_1)
\\a_t\sim \pi\left(a_t|z_t \right )
\\z_{t+1}\sim p\left(z_{t+1}|z_t,a_t \right )}}{\mathbb{E}}
\sum_{t=1}^{H}r\left(z_t,a_t \right )
\end{equation}

Note here we do not explicitly write the discount factor $\gamma$ in the accumulative rewards, instead we incorporate the discount factor by modifying the state transition model~\cite{levine2018reinforcement}. If the initial state transitions are given by $p\left ( z_{t+1}|z_t, a_t \right )$, adding a discount factor is equivalent to undiscounted problem under the modified state transitions $\bar{p}\left ( z_{t+1}|z_t, a_t \right )=\gamma\,p\left ( z_{t+1}|z_t, a_t \right )$, where there is an additional transition with probability $1-\gamma$, regardless of action, into an absorbing state with reward zero. The discount factor allows convergence of the value function in infinite-horizon settings. Without loss of generality, we will omit $\gamma$ from the PGM related derivations in this paper, but it can be inserted trivially in all cases simply by modifying the state transition models as mentioned above. The discount factor is revisited as an explicit consideration in our reinforcement learning algorithm implementation in \ref{sec:policylearning}.

Maximum entropy reinforcement learning (MaxEnt RL)~\cite{haarnoja2017reinforcement,haarnoja2018soft,levine2018reinforcement} modifies the above standard RL by adding an entropy regularization term ${\cal H}\left(\pi\left(a_t|z_t \right ) \right )=-\textrm{log}\pi\left(a_t|z_t \right )$ to the reward. Now considering we are using a parametric function as the policy $\pi_\phi$, for example a deep neural network with weights $\phi$, then the objective of MaxEnt RL can be written as:
\begin{equation}
    \phi^* =\underset{\phi}{\textrm{argmax}}\underset{\substack{z_1 \sim p(z_1)
\\a_t\sim \pi_\phi \left(a_t|z_t \right )
\\z_{t+1}\sim p\left(z_{t+1}|z_t,a_t \right )}}{\mathbb{E}}
\sum_{t=1}^{H}\left[r\left(z_t,a_t \right ) -\textrm{log}\pi_\phi\left(a_t|z_t \right ) \right ]
\end{equation}

There are several reasons why we would like to use MaxEnt RL instead of standard RL~\cite{eysenbach2019if}. First, it performs better exploration. Standard RL requires specific exploration strategies such as adding noise to the policy. However, MaxEnt RL has a stochastic policy by default, thus the policy itself includes the exploration strategy, which is optimized during RL training. In practice, the performance of MaxEnt RL is often better and more robust than standard RL algorithms. 

\begin{figure}
  \centering
  \includegraphics[width = .45\textwidth]{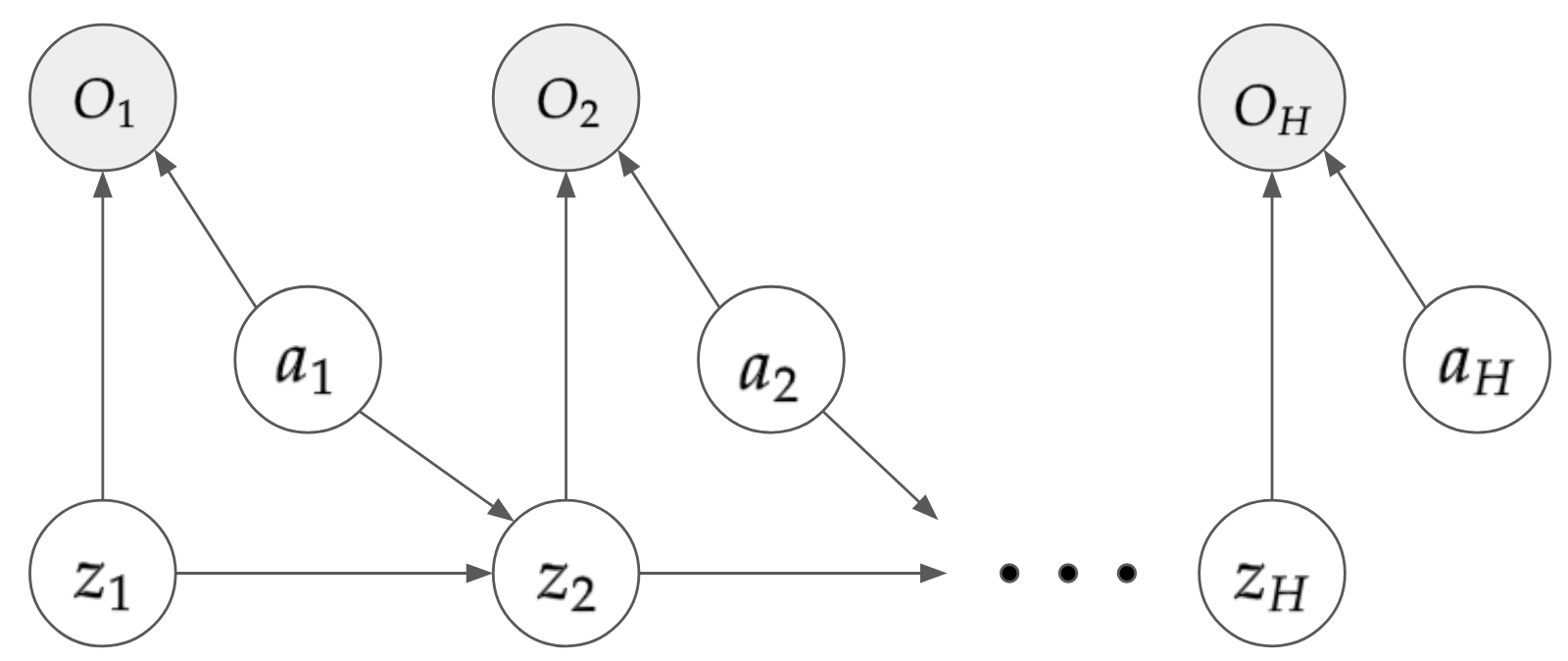}
  \caption{\label{Fig:PGMRL}A PGM for maximum entropy reinforcement learning.}
\end{figure}

Second, MaxEnt RL can be interpreted as learning a PGM. As shown in Fig.\ref{Fig:PGMRL}, $z_t$ represents the state, $a_t$ is the action, and $O_t$ is a binary random variable. The use of $O_t$ is to indicate whether the agent is acting optimally at time step t. Its conditional probability is defined by:
\begin{equation}
    p\left(O_t=1|z_t,a_t \right )=\textrm{exp}\left(r\left(z_t,a_t \right ) \right )
\end{equation}
thus higher reward indicates higher optimality. Therefore, to make the agent act optimally, we want to maximize the probability of optimality in the whole trajectory $p\left(O_{1:H} \right )$. Let's now look at its log probability:
\begin{equation}\label{Eq:MaxEntRL}
\begin{aligned}
\textrm{log}\,p\left(O_{1:H} \right )
&=\textrm{log}\int\int p\left(O_{1:H}, z_{1:H}, a_{1:H} \right )dz_{1:H}da_{1:H}\\
&= \textrm{log}\int\int p\left(O_{1:H}, z_{1:H}, a_{1:H} \right )\\
&\qquad \qquad \qquad \frac{q\left(z_{1:H},a_{1:H}\right)}{q\left(z_{1:H},a_{1:H}\right)}dz_{1:H}da_{1:H}\\
&=\textrm{log}\,\underset{q\left(z_{1:H},a_{1:H} \right)}{\mathbb{E}}\left[\frac{p\left(O_{1:H}, z_{1:H}, a_{1:H} \right )}{q\left(z_{1:H},a_{1:H}\right)} \right]\\
&\geq \underset{q\left(z_{1:H},a_{1:H} \right)}{\mathbb{E}}\,[\textrm{log}\, p\left(O_{1:H}, z_{1:H}, a_{1:H} \right )\\
&\qquad \qquad -\textrm{log}\,q\left(z_{1:H},a_{1:H} \right)]
\end{aligned}
\end{equation}

The above inequality is obtained by adding a variational distribution $q\left(z_{1:H},a_{1:H} \right)$ and then applying Jensen's inequality. The variational distribution should be the trajectory distribution generated by the current policy $\pi\left(a_t|z_t\right)$:
\begin{equation}
    q\left(z_{1:H},a_{1:H}\right)=p\left(z_1\right)\pi\left(a_H|z_H\right)\prod_{t=1}^{H-1}p\left(z_{t+1}|z_t,a_t\right)\pi\left(a_t|z_t\right)
\end{equation}

The optimality distribution of the trajectory is:
\begin{equation}
\begin{aligned}
p\left(O_{1:H}, z_{1:H}, a_{1:H} \right )
&=p\left(O_{1:H}|z_{1:H}, a_{1:H} \right )p\left(z_{1:H}, a_{1:H} \right )\\
&=\textrm{exp}\left(\sum_{t=1}^Hr\left(z_t,a_t\right)\right)\\
&\qquad \qquad p\left(z_1\right)\prod_{t=1}^{H-1}p\left(z_{t+1}|z_t,a_t\right)
\end{aligned}
\end{equation}

By cancellation of repeated terms, the inequality \eqref{Eq:MaxEntRL} becomes:
\begin{equation}
\textrm{log}\,p\left(O_{1:H}\right) \geq \underset{q\left(z_{1:H},a_{1:H}\right)}{\mathbb{E}}\sum_{t=1}^H \left[r\left(z_t,a_t\right)-\textrm{log}\,\pi\left(a_t|z_t\right)\right]    
\end{equation}

Note that we can maximize the left side by maximizing the right side, and the right side of the inequality is exactly the same objective of MaxEnt RL. This means, we can use MaxEnt RL to maximize the likelihood of optimality variables in the PGM in Fig.\ref{Fig:PGMRL}. In this sense, the reinforcement learning problem is reformulated into a learning problem for the PGM shown in Fig.\ref{Fig:PGMRL}.

\section{Interpretable End-to-end Urban Autonomous Driving}
\subsection{PGM for Interpretable Urban Autonomous Driving}
There are two main building blocks for urban autonomous driving. The first is the perception and recognition module, which helps the autonomous car to understand the current driving situation, such as where is the ego vehicle, what is the road condition, and where are the surrounding road participants. Furthermore, it needs to be able to reason about what will happen in the future, such as where will the ego car and surrounding road participants go. These information should be obtained given the historical high dimensional raw sensor inputs. The second module is planning and control, which helps the autonomous car decide what action to take. 

\begin{figure}
  \centering
  \includegraphics[width = .48\textwidth]{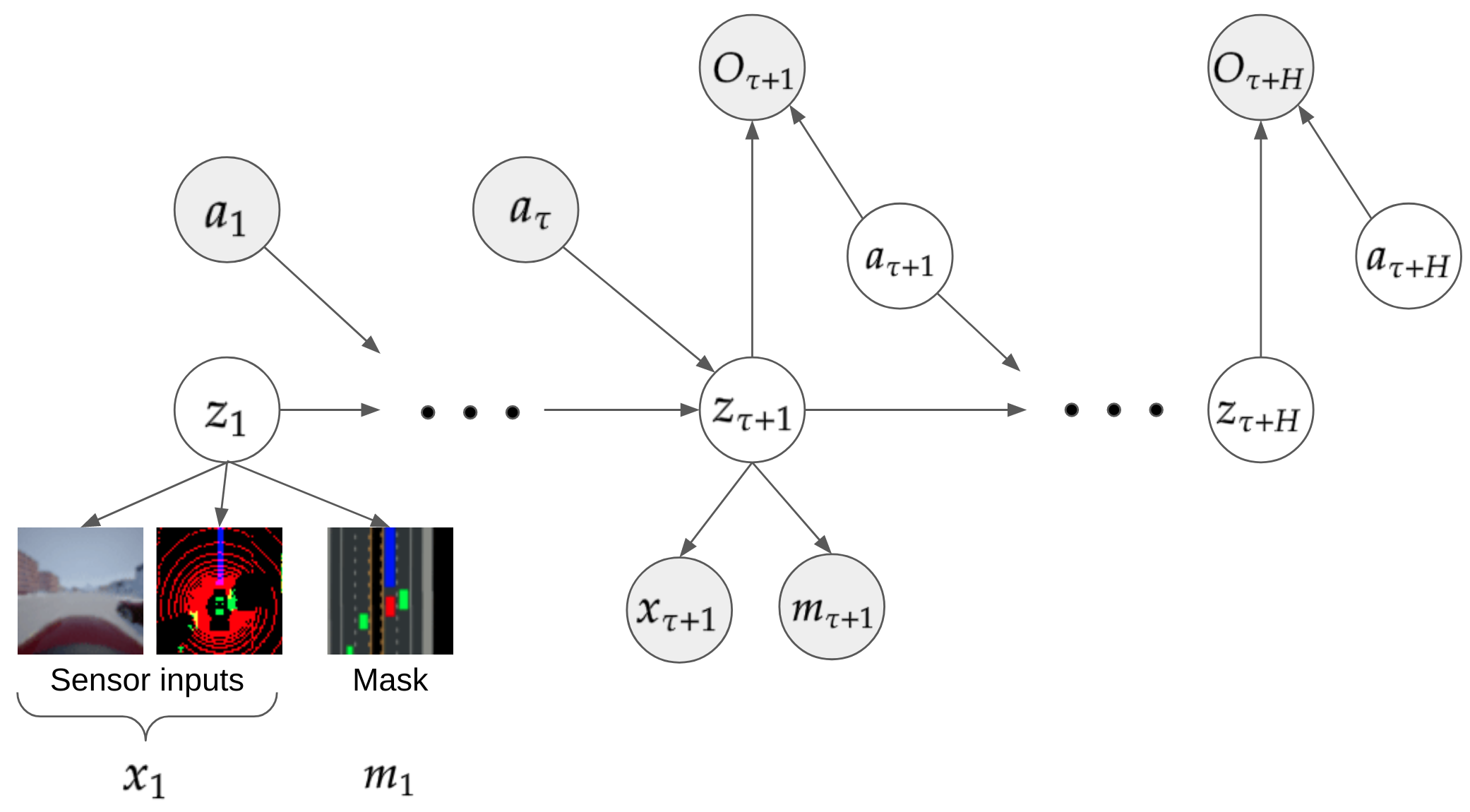}
  \caption{\label{Fig:PGMdriving}A PGM for interpretable end-to-end urban autonomous driving}
\end{figure}

Using the methods mentioned in Section~\ref{sec:PGM}, the above two building blocks can be formulated by two PGMs separately, and it's natural to combine the two PGMs into a single one. Inspired by recent works that combines latent representation learning and reinforcement learning~\cite{lee2019stochastic,hafner2018learning,ha2018world}, we present our PGM for urban autonomous driving, as shown in Fig.\ref{Fig:PGMdriving}. Same to the notations in Section~\ref{sec:PGM}, $z_t$ represents for the latent state, $a_t$ represents for action, $O_t$ represents for the optimality variable, and $x_t$ represents for the sensor inputs. Note here we allow sensor inputs from multiple sources.

We have a newly introduced variable, $m_t$, which we call the mask. It contains semantic meanings of the environment in a human understandable way. Details about this mask is described in Section~\ref{sec:mask}. The main purpose of the mask is to provide interpretability for the system. At training time we need to provide the ground truth labels of the mask, but at test time, the mask can be decoded from the latent state, showing how the system is understanding the environment semantically.

After learning this PGM in Fig.\ref{Fig:PGMdriving}, the following modules can be obtained:
\subsubsection{\bf{Policy $p\left(a_t|z_t\right)$}}
Given the latent state, the policy model tells how to choose the action.
\subsubsection{\bf{Inference $p\left(z_{t+1}|x_{1:t+1},a_{1:t}\right)$}}
With historical sensor inputs and actions, the inference model infers the current latent state.
\subsubsection{\bf{Latent dynamics $p\left(z_{t+1}|z_t,a_t\right)$}}
This helps predict the future states.
\subsubsection{\bf{Generative models $p\left(x_t|z_t\right),\,p\left(m_t|z_t\right)$}}
$p\left(x_t|z_t\right)$ decodes the latent state $z_t$ to raw sensor inputs $x_t$, showing how much information the latent state captures. $p\left(m_t|z_t\right)$ generates the semantic mask $m_t$ to provide interpretability.

\vspace{1mm}

\begin{figure}
  \centering
  \includegraphics[width = .48\textwidth]{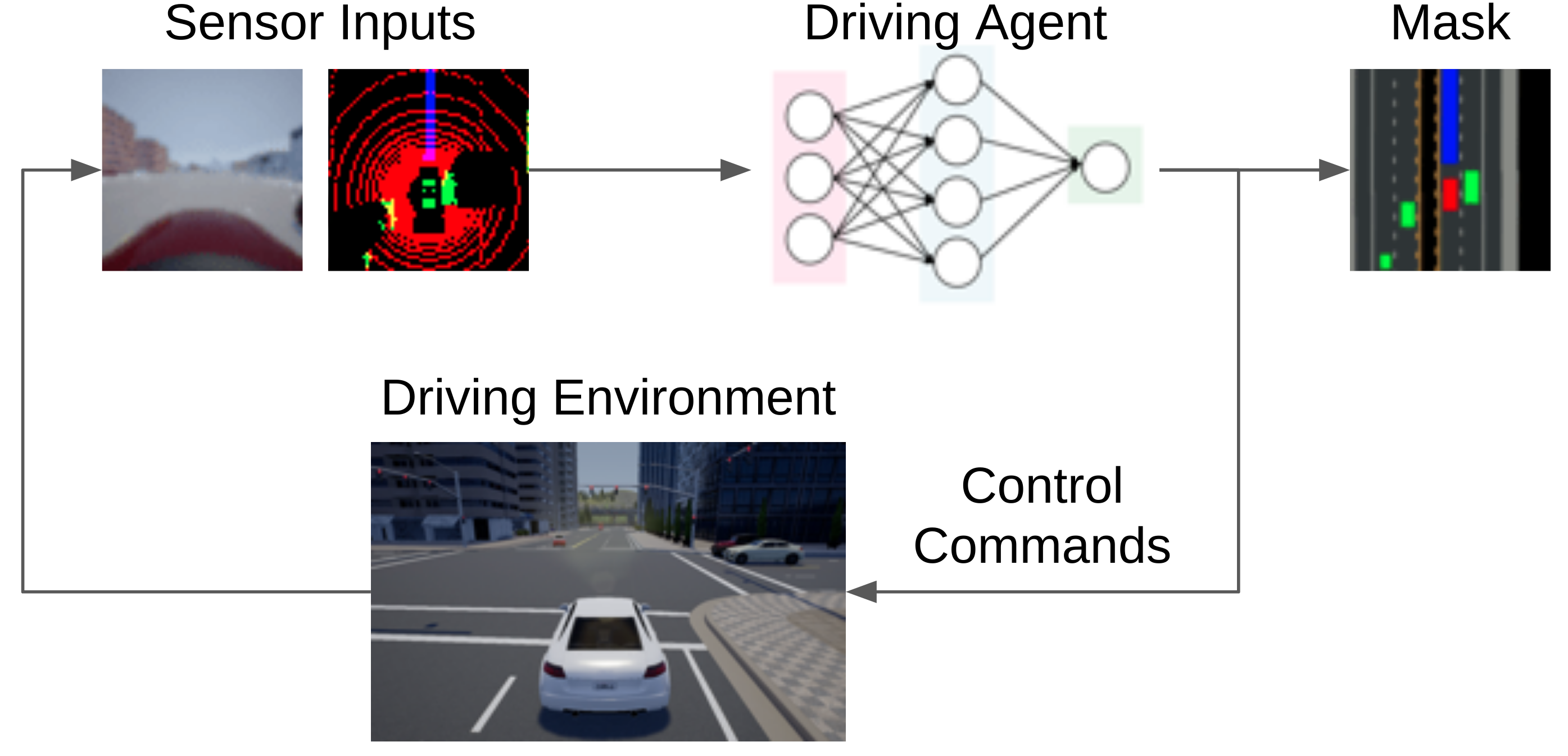}
  \caption{\label{Fig:agent}The interpretable end-to-end urban autonomous driving agent}
\end{figure}

The whole model can be trained end-to-end. After training, an intelligent driving agent containing an interpretable environment model and a driving policy is obtained. As shown in Fig.\ref{Fig:agent}, the agent takes multi-modal sensor inputs from the driving environment, and then output control commands to drive the car in urban scenarios. In the meantime, the agent generates a semantic mask to interpret how it understand the current driving situation.

\subsection{Sensor Inputs and Mask}\label{sec:mask}
We use two sensors to provide the observations, camera and lidar. For camera, the sensor input is a front-view RGB image, which can be represented by a tensor of $\mathbb{R}^{64\times 64\times 3}$. For lidar, we project the point clouds to the ground plane and render them into a 2D lidar image. The lidar image is represented by a tensor of $\mathbb{R}^{64\times 64\times 3}$, with each pixel rendered in red or green depending on whether there are lidar points at or above ground level existing in the corresponding pixel cell. Desired route constituted of waypoints are rendered in blue.

We use camera and lidar together because they are both important sensor sources and provide complementary information. Lidar point clouds provides accurate spatial information of other road participants and obstacles in 360 degrees of view. While the front-view camera is good at providing information of the road conditions. 

\begin{figure}
  \centering
  \includegraphics[width = .45\textwidth]{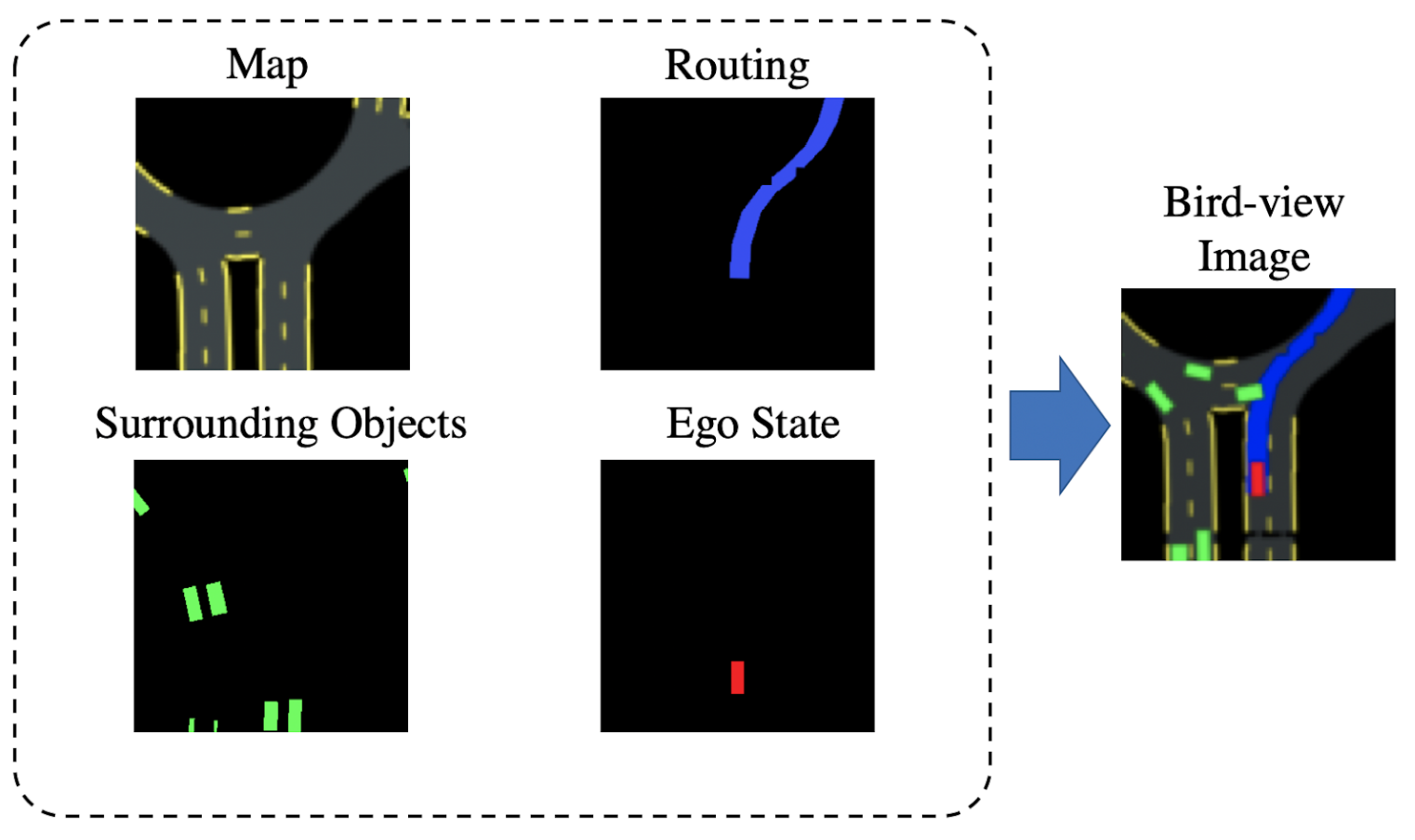}
  \caption{\label{Fig:mask}The bird-view semantic mask for urban autonomous driving.}
\end{figure}

The semantic mask provides bird-view semantics of the road conditions and objects, which is represented by a tensor of $\mathbb{R}^{64\times 64\times 3}$. As shown in Fig.\ref{Fig:mask}, the mask is composed of the following four parts:

\subsubsection{\bf{Map}}
Map contains information of road conditions. Drivable areas and lane markings are rendered in the map.

\subsubsection{\bf{Routing}}
Routing contains information of waypoints, which is provided by a route planner. It is rendered as a thick blue polyline.

\subsubsection{\bf{Detected Objects}}
Historical bounding boxes of detected surrounding road participants (e.g, vehicles, bicycles and pedestrians) are rendered as green boxes.

\subsubsection{\bf{Ego State}}
The bounding box of the ego vehicle is rendered as a red box.

\section{Joint Learning of Environment Model and Driving Policy}
\subsection{Variational Inference for Joint Model Learning and Policy Learning}\label{Sec:variational}
The environment model and driving policy can be learned jointly by learning the PGM shown in Fig.\ref{Fig:PGMdriving}. For convenience, we first introduce some notations. Denote a trajectory to be composed of sensor inputs, masks, actions and rewards:
\begin{equation}
    \vec{x}=x_{1:\tau+1},\,\vec{m}=m_{1:\tau+1},\,\vec{a}=a_{1:\tau},\,\vec{r}=r_{1:\tau}
\end{equation}

The dataset is then composed of this kind of trajectories collected during the exploration phase ${\cal D} = \left \{ \left(\vec{x}^i, \vec{m}^i, \vec{a}^i, \vec{r}^i \right ) \right \}_{i=1}^N$. We further denote:
\begin{equation}
\begin{aligned}
&\vec{z}=z_{1:\tau+1},\,\vec{z}^w=z_{1:\tau+H},\,\vec{z}^p=z_{\tau+1:\tau+H},\\
&\vec{O}^p=O_{\tau+1:\tau+H},\,\vec{a}^p=a_{\tau+1:\tau+H}
\end{aligned}
\end{equation}
where the superscript “p” stands for “post”, and “w” stands for “whole”. The learning objective is to maximize the log likelihood of the sensor inputs, mask and the optimality variables:
\begin{equation}
\textrm{log}\prod_{\left(\vec{x}, \vec{m}, \vec{a}, \vec{r} \right ) \in \cal D} p\left(\vec{x},\vec{m},\vec{O}^p|\vec{a} \right)=\sum_{\left(\vec{x}, \vec{m}, \vec{a}, \vec{r} \right ) \in \cal D}\textrm{log}\,p\left(\vec{x},\vec{m},\vec{O}^p|\vec{a} \right)
\end{equation}

This can be maximized by stochastic gradient descent (SGD), which optimizes parametric functions by gradient descent, with the gradient estimated by sampling a batch of data points. To make SGD applicable to our problem, $p\left(\vec{x},\vec{m},\vec{O}^p|\vec{a} \right)$ needs to be represented by parametric functions, then auto-differentiation tools (e.g, TensorFlow) can be used to calculate its gradient. We can use variational inference~\cite{kingma2013auto} to compute this log likelihood. We first introduce the latent variables $\vec{z}^w$ and $\vec{a}^p$:
\begin{equation}\label{Eq:latent}
\begin{aligned}
\textrm{log}\,p\left(\vec{x},\vec{m},\vec{O}^p|\vec{a} \right)=\textrm{log}\int \int p\left(\vec{x},\vec{m},\vec{O}^p,\vec{z}^w,\vec{a}^p|\vec{a} \right)d \vec{z}^w d \vec{a}^p
\end{aligned}
\end{equation}

Then introduce a variational distribution $q\left(\vec{z}^w,\vec{a}^p|\vec{x},\vec{a}\right)$ into \eqref{Eq:latent}:
\begin{equation}\label{Eq:vardist}
\begin{aligned}
&\textrm{log}\,p\left(\vec{x},\vec{m},\vec{O}^p|\vec{a} \right)\\
&=\textrm{log}\int \int p\left(\vec{x},\vec{m},\vec{O}^p,\vec{z}^w,\vec{a}^p|\vec{a} \right)\frac{q\left(\vec{z}^w,\vec{a}^p|\vec{x},\vec{a}\right)}{q\left(\vec{z}^w,\vec{a}^p|\vec{x},\vec{a}\right)} d \vec{z}^w d \vec{a}^p
\end{aligned}
\end{equation}

The variational distribution is defined as:
\begin{equation}\label{Eq:variational}
\begin{aligned}
&q\left(\vec{z}^w,\vec{a}^p|\vec{x},\vec{a}\right)\\
&=q\left(\vec{z}|\vec{x},\vec{a}\right)\pi\left(a_{\tau+H}|z_{\tau+H}\right)\prod_{t=\tau+1}^{\tau+H-1}p\left(z_{t+1}|z_t,a_t\right)\pi\left(a_t|z_t\right)
\end{aligned}
\end{equation}
where $q\left(\vec{z}|\vec{x},\vec{a}\right)$ is the inference of latent states given historical sensor inputs and actions. The rest part of the right hand side represents the trajectory distribution by executing policy $\pi\left(a_t|z_t\right)$ with latent state transition $p\left(z_{t+1}|z_t,a_t\right)$.

Now eliminate the integration in \eqref{Eq:vardist} by introducing expectation, and apply Jensen's inequality we have:
\begin{equation}\label{Eq:ELBO}
\begin{aligned}
\textrm{log}\,p\left(\vec{x},\vec{m},\vec{O}^p|\vec{a} \right)
&=\textrm{log}\underset{q\left(\vec{z}^w,\vec{a}^p|\vec{x},\vec{a}\right)}{\mathbb{E}}\left[\frac{p\left(\vec{x},\vec{m},\vec{O}^p,\vec{z}^w,\vec{a}^p|\vec{a} \right)}{q\left(\vec{z}^w,\vec{a}^p|\vec{x},\vec{a}\right)}\right]\\
&\geq \underset{q\left(\vec{z}^w,\vec{a}^p|\vec{x},\vec{a}\right)}{\mathbb{E}}\left[\textrm{log}\,p\left(\vec{x},\vec{m},\vec{O}^p,\vec{z}^w,\vec{a}^p|\vec{a} \right)\right.\\
&\qquad \qquad \qquad \qquad \left.-\textrm{log}\,q\left(\vec{z}^w,\vec{a}^p|\vec{x},\vec{a}\right)\right]\\
&=\text{ELBO}
\end{aligned}
\end{equation}
where ELBO stands for evidence lower bound. We can maximize the original log likelihood by maximizing the ELBO. Let's now derive $p\left(\vec{x},\vec{m},\vec{O}^p,\vec{z}^w,\vec{a}^p|\vec{a} \right)$ by probability factorization according to the PGM in Fig.\ref{Fig:PGMdriving}:
\begin{equation}
\begin{aligned}
&p\left(\vec{x},\vec{m},\vec{O}^p,\vec{z}^w,\vec{a}^p|\vec{a} \right)\\
&\qquad \qquad=p\left(\vec{x},\vec{m},\vec{O}^p,z_{\tau+2:\tau+H},\vec{a}^p|\vec{z},\vec{a} \right)p\left(\vec{z}|\vec{a}\right)\\
&\qquad \qquad=p\left(\vec{x}|\vec{z}\right)p\left(\vec{m}|\vec{z}\right)p\left(\vec{O}^p,z_{\tau+2:\tau+H},\vec{a}^p|z_{\tau+1} \right)p\left(\vec{z}|\vec{a}\right)\\
&\qquad \qquad=p\left(\vec{x}|\vec{z}\right)p\left(\vec{m}|\vec{z}\right)\frac{p\left(\vec{O}^p,\vec{z}^p,\vec{a}^p \right)}{p\left(z_{\tau+1}\right)}p\left(\vec{z}|\vec{a}\right)
\end{aligned}
\end{equation}

According to the soft optimality assumption:
\begin{equation}
\begin{aligned}
&p\left(\vec{O}^p,\vec{z}^p,\vec{a}^p \right)=p\left(\vec{z}^p,\vec{a}^p \right)p\left(\vec{O}^p|\vec{z}^p,\vec{a}^p \right)\\
&=p\left(\vec{a}^p\right)p\left(z_{\tau+1}\right)\prod_{t=\tau+1}^{\tau+H-1}p\left(z_{t+1}|z_t,a_t\right)\textrm{exp}\left(\sum_{t=\tau+1}^{\tau+H}r\left(z_t,a_t\right)\right)
\end{aligned}
\end{equation}

We thus have:
\begin{equation}
\begin{aligned}
&p\left(\vec{x},\vec{m},\vec{O}^p,\vec{z}^w,\vec{a}^p|\vec{a} \right)=p\left(\vec{x}|\vec{z}\right)p\left(\vec{m}|\vec{z}\right)p\left(\vec{a}^p\right)\\
&\qquad \qquad \prod_{t=\tau+1}^{\tau+H-1}p\left(z_{t+1}|z_t,a_t\right)\textrm{exp}\left(\sum_{t=\tau+1}^{\tau+H}r\left(z_t,a_t\right)\right)p\left(\vec{z}|\vec{a}\right)
\end{aligned}
\end{equation}

Substituting the variational distribution \eqref{Eq:variational} into \eqref{Eq:ELBO}, we have:
\begin{equation}\label{Eq:ELBO2}
\begin{aligned}
&\textrm{ELBO}=\underset{q\left(\vec{z}^w,\vec{a}^p|\vec{x},\vec{a}\right)}{\mathbb{E}}\left[\textrm{log}\,p\left(\vec{x}|\vec{z}\right)+\textrm{log}\,p\left(\vec{m}|\vec{z}\right)+\textrm{log}\,p\left(\vec{z}|\vec{a}\right)\right.\\
&+\textrm{log}\prod_{t=\tau+1}^{\tau+H-1}p\left(z_{t+1}|z_t,a_t\right)+\sum_{t=\tau+1}^{\tau+H}r\left(z_t,a_t\right)-\textrm{log}\,q\left(\vec{z}|\vec{x},\vec{a}\right)\\
&\left.-\textrm{log}\prod_{t=\tau+1}^{\tau+H}\pi\left(a_t|z_t\right)-\textrm{log}\prod_{t=\tau+1}^{\tau+H-1}p\left(z_{t+1}|z_t,a_t\right)+\text{log}\,p\left(\vec{a}^p\right)\right]
\end{aligned}
\end{equation}

Notice the cancellations in \eqref{Eq:ELBO2}, we have:
\begin{equation}\label{Eq:ELBO3}
\begin{aligned}
&\textrm{ELBO}=\underset{q\left(\vec{z}^w,\vec{a}^p|\vec{x},\vec{a}\right)}{\mathbb{E}}\left[\textrm{log}\,p\left(\vec{x}|\vec{z}\right)+\textrm{log}\,p\left(\vec{m}|\vec{z}\right)+\textrm{log}\,p\left(\vec{z}|\vec{a}\right)\right.\\
&\qquad \qquad \qquad \qquad \qquad \qquad \left.-\textrm{log}\,q\left(\vec{z}|\vec{x},\vec{a}\right)\right]\\
&+\underset{q\left(\vec{z}^w,\vec{a}^p|\vec{x},\vec{a}\right)}{\mathbb{E}}\left[\sum_{t=\tau+1}^{\tau+H}\left(r\left(z_t,a_t\right)-\textrm{log}\,\pi\left(a_t|z_t\right)+\text{log}\,p\left(a_t\right)\right)\right]
\end{aligned}
\end{equation}

The first part of the right hand side of \eqref{Eq:ELBO3} corresponds to learning the environment model, while the second part corresponds to learning the driving policy, we will derive the details of the two parts in \ref{sec:modellearning} and \ref{sec:policylearning}, respectively.

\subsection{Environment Model Learning}\label{sec:modellearning}
The environment model can be learned via optimizing the first part of \eqref{Eq:ELBO3}:
\begin{equation}\label{Eq:modellearning}
\begin{aligned}
\underset{q\left(\vec{z}|\vec{x},\vec{a}\right)}{\mathbb{E}}\left[\textrm{log}\,p\left(\vec{x}|\vec{z}\right)+\textrm{log}\,p\left(\vec{m}|\vec{z}\right)+\textrm{log}\,p\left(\vec{z}|\vec{a}\right)-\textrm{log}\,q\left(\vec{z}|\vec{x},\vec{a}\right)\right]
\end{aligned}
\end{equation}
where we replace $\mathbb{E}_{q\left(\vec{z}^w,\vec{a}^p|\vec{x},\vec{a}\right)}$ with $\mathbb{E}_{q\left(\vec{z}|\vec{x},\vec{a}\right)}$ because this part of ELBO is only related to $z_{1:\tau+1}$. Now let's further derive the components in \eqref{Eq:modellearning} by unfolding them with time. Considering the conditional dependence of PGM in Fig.\ref{Fig:PGMdriving}. The generative models can be unfolded as:
\begin{equation}
\begin{aligned}
&\textrm{log}\,p\left(\vec{x}|\vec{z}\right)=\textrm{log}\prod_{t=1}^{\tau+1}p\left(x_t|z_t\right)=\sum_{t=1}^{\tau+1}\textrm{log}\,p\left(x_t|z_t\right)\\
&\textrm{log}\,p\left(\vec{m}|\vec{z}\right)=\textrm{log}\prod_{t=1}^{\tau+1}p\left(m_t|z_t\right)=\sum_{t=1}^{\tau+1}\textrm{log}\,p\left(m_t|z_t\right)
\end{aligned}
\end{equation}

The prior model can be unfolded using the latent state transition function:
\begin{equation}
\begin{aligned}
\textrm{log}\,p\left(\vec{z}|\vec{a}\right)
&=\textrm{log}\left[p\left(z_1\right)\prod_{t=1}^\tau p\left(z_{t+1}|z_t,a_t\right)\right]\\
&=\textrm{log}\,p\left(z_1\right)+\sum_{t=1}^\tau \textrm{log}\,p\left(z_{t+1}|z_t,a_t\right)
\end{aligned}
\end{equation}

The posterior inference model can be unfolded as:
\begin{equation}
\begin{aligned}
\textrm{log}\,q\left(\vec{z}|\vec{x},\vec{a}\right)
&=\textrm{log}\left[q\left(z_1|\vec{x},\vec{a}\right)\prod_{t=1}^\tau q\left(z_{t+1}|z_t,\vec{x},\vec{a}\right)\right]\\
&\approx \textrm{log}\left[q\left(z_1|x_1\right)\prod_{t=1}^\tau q\left(z_{t+1}|z_t,x_{t+1},a_t\right)\right]\\
&=\textrm{log}\,q\left(z_1|x_1\right)+\sum_{t=1}^\tau\textrm{log}\, q\left(z_{t+1}|z_t,x_{t+1},a_t\right)
\end{aligned}
\end{equation}

Note here we approximate $q\left(\vec{z}|\vec{x},\vec{a}\right)$ and $q\left(z_{t+1}|z_t,\vec{x},\vec{a}\right)$ with $q\left(z_1|x_1\right)$ and $q\left(z_{t+1}|z_t,x_{t+1},a_t\right)$ for simplicity. If we want to obtain the exact accurate values, bi-directional recurrent neural networks should be used to obtain the posterior probabilities conditioned on the whole trajectory sequence $\left(\vec{x},\,\vec{a}\right)$~\cite{krishnan2015deep}.

We can now unfold \eqref{Eq:modellearning} with time:
\begin{equation}\label{Eq:modelELBO}
\begin{aligned}
&\underset{q\left(\vec{z}|\vec{x},\vec{a}\right)}{\mathbb{E}}\left[\textrm{log}\,p\left(\vec{x}|\vec{z}\right)+\textrm{log}\,p\left(\vec{m}|\vec{z}\right)+\textrm{log}\,p\left(\vec{z}|\vec{a}\right)-\textrm{log}\,q\left(\vec{z}|\vec{x},\vec{a}\right)\right]\\
&\qquad \qquad \approx \underset{q\left(\vec{z}|\vec{x},\vec{a}\right)}{\mathbb{E}}\left[\sum_{t=1}^{\tau+1}\textrm{log}\,p\left(x_t|z_t\right)+\sum_{t=1}^{\tau+1}\textrm{log}\,p\left(m_t|z_t\right)\right.\\
&\qquad \qquad-\text{D}_{\text{KL}}\left(q\left(z_1|x_1\right)||p\left(z_1\right)\right)\\
&\qquad \qquad \left.-\sum_{t=1}^{\tau+1}\text{D}_{\text{KL}}\left(q\left(z_{t+1}|z_t,x_{t+1},a_t\right)||p\left(z_{t+1}|z_t,a_t\right)\right)\right]
\end{aligned}
\end{equation}

\subsection{Driving Policy Learning}\label{sec:policylearning}
The driving policy can be learned via optimizing the second part of \eqref{Eq:ELBO3}:
\begin{equation}\label{Eq:policylearning}
\begin{aligned}
&\text{max}\,\underset{q\left(\vec{z}^p,\vec{a}^p|\vec{x},\vec{a}\right)}{\mathbb{E}}\sum_{t=\tau+1}^{\tau+H}\left[r\left(z_t,a_t \right ) -\textrm{log}\pi_\phi\left(a_t|z_t \right ) +\text{log}\,p\left(a_t\right) \right ]\\
&=\underset{\substack{z_{\tau+1} \sim p\left(z_{\tau+1}|\vec{x},\vec{a}\right)
\\a_t\sim \pi_\phi \left(a_t|z_t \right )
\\z_{t+1}\sim p\left(z_{t+1}|z_t,a_t \right )}}{\mathbb{E}}
\sum_{t=\tau+1}^{\tau+H}\left[r\left(z_t,a_t \right ) -\textrm{log}\,\pi_\phi\left(a_t|z_t \right ) \right ]
\end{aligned}
\end{equation}
where $\text{log}\,p\left(a_t\right)$ is ignored since we assume uniform action prior. The optimization problem \eqref{Eq:policylearning} then becomes a standard MaxEnt RL problem. 

We use soft actor-critic (SAC)~\cite{haarnoja2018soft} to solve this MaxEnt RL problem. SAC is a function approximation version of the soft policy iteration (SPI). SPI is an extension of the standard policy iteration to the maximum entropy case, which is to iteratively apply the soft policy evaluation:
\begin{equation}\label{Eq:SPE}
\begin{aligned}
&{\cal T}^\pi Q\left(z_t,a_t\right)=r\left(z_t,a_t\right)\\
&\qquad \; +\gamma\,\underset{z_{t+1} \sim p}{\mathbb{E}}\left[\underset{a_{t+1} \sim \pi}{\mathbb{E}}\left[Q\left(z_{t+1},a_{t+1}\right)-\text{log}\,\pi\left(a_{t+1}|z_{t+1}\right)\right]\right] 
\end{aligned}
\end{equation}
and the soft policy improvement:
\begin{equation}\label{Eq:SPI}
\pi_{\textrm{new}}=\underset{\pi'}{\textrm{argmin}}\,\text{D}_{\text{KL}}\left(\pi'\left(\cdot|z_t\right)\left|\left|\frac{\textrm{exp}\left(Q^{\pi_{\textrm{old}}}\left(z_t,\cdot\right)\right)}{Z^{\pi_{\textrm{old}}}\left(z_t\right)}\right.\right.\right)
\end{equation}
where $Z^{\pi_{\textrm{old}}}\left(z_t\right)$ is the normalization term.

The function approximation implementation is to optimize the loss functions that address the soft policy evaluation and soft policy improvement. The loss functions are the Bellman residual in \eqref{Eq:SPE}:
\begin{equation}\label{Eq:JQ}
J_Q=\underset{z_\tau \sim q\left(\vec{z}|\vec{x},\vec{a}\right)}{\mathbb{E}}\left[\frac{1}{2}\left(Q\left(z_\tau,a_\tau\right)-\hat{Q}\left(z_\tau,a_\tau\right)\right)\right]
\end{equation}
and the KL divergence in \eqref{Eq:SPI}:
\begin{equation}\label{Eq:Jpi}
J_\pi=\underset{\substack{z_{\tau+1} \sim q\left(\vec{z}|\vec{x},\vec{a}\right)\\
a_{\tau+1} \sim \pi\left(a_{\tau+1}|z_{\tau+1}\right)}}{\mathbb{E}}\left[\textrm{log}\,\pi\left(a_{\tau+1}|z_{\tau+1}\right)-Q\left(z_{\tau+1},a_{\tau+1}\right)\right]
\end{equation}

Note
\begin{equation}
\begin{aligned}
&\hat{Q}\left(z_\tau,a_\tau\right)=r_\tau\\
&\quad+\gamma\,\underset{\substack{z_{\tau+1} \sim q\left(\vec{z}|\vec{x},\vec{a}\right)\\
a_{\tau+1} \sim \pi\left(a_{\tau+1}|z_{\tau+1}\right)}}{\mathbb{E}}\left[\Bar{Q}\left(z_{\tau+1},a_{\tau+1}\right)-\textrm{log}\,\pi\left(a_{\tau+1}|z_{\tau+1}\right)\right]
\end{aligned}
\end{equation}
where $\bar{Q}$ is a delayed Q network.

Thus, the joint learning algorithm becomes to use SGD to maximize the model learning part of ELBO in \eqref{Eq:modelELBO} and minimize $J_Q$ in \eqref{Eq:JQ} and $J_\pi$ in \eqref{Eq:Jpi}.

\section{Experiments}
\subsection{Simulation Setup}
\begin{figure}
    \centering
    \begin{subfigure}[b]{0.22\textwidth}
        \centering
        \includegraphics[height = \linewidth]{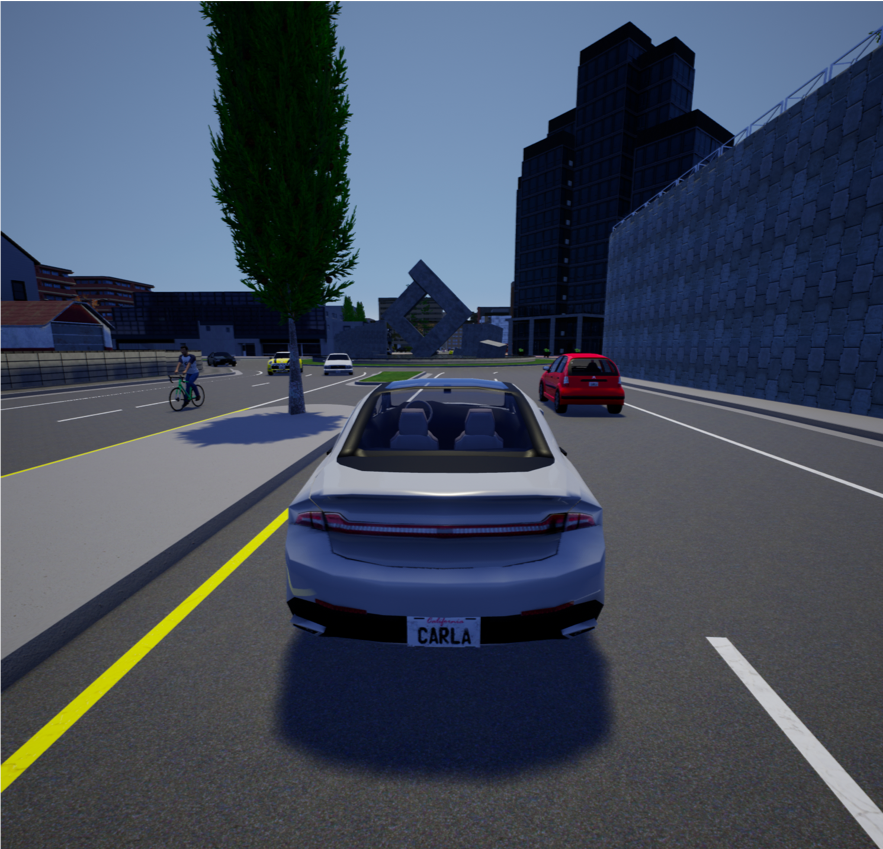}
        \caption{Sample view of CARLA simulator}
    \end{subfigure}
    ~~
    \begin{subfigure}[b]{0.22\textwidth}
        \centering
        \includegraphics[height = \linewidth]{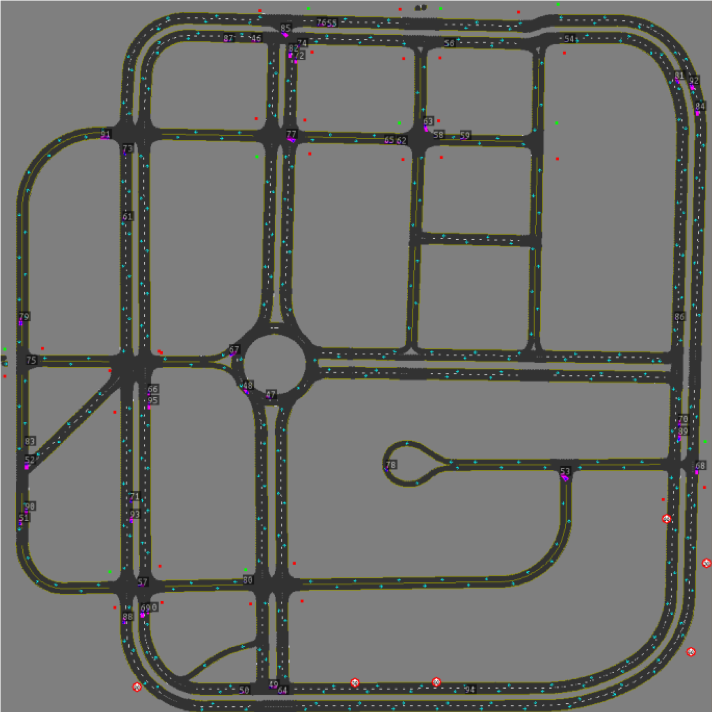}
        \caption{Map layout of the simulated city}
    \end{subfigure}
    \caption{\label{Fig:carla}Simulation environment}
\end{figure}

We train and evaluate our proposed method on CARLA simulator~\cite{dosovitskiy2017carla}. CARLA is a high-definition open-source simulation platform for autonomous driving research. It simulates not only the driving environment and vehicle dynamics, but also the raw sensor data inputs such as camera RGB image and lidar point cloud. Fig.\ref{Fig:carla} (a) shows a sample view of the driving simulation environment we use.

Fig.\ref{Fig:carla} (b) shows the map layout of the virtual town in CARLA we use for training. It includes various urban scenarios such as intersections and roundabouts. The range of the map is $400m \times 400m$, with about $6km$ total length of roads. 100 vehicles are running autonomously in the virtual town to simulate a multi-agent environment. The vehicles will randomly choose a direction at intersections, then follow the route, while slowing down for front vehicles and stopping when the front traffic light becomes red.

\subsection{Implementation Details}
\subsubsection{Reward Function}
We use the following reward function in our experiments:
\begin{equation}\label{Eq:reward}
r = 200\,r_{\textrm{collision}} + v_{\textrm{lon}} + 10\,r_{\textrm{fast}} + r_{\textrm{out}} - 5\,\alpha^2 + 0.2\,r_{\textrm{lat}} - 0.1
\end{equation}
where $r_{\textrm{collision}}$ is the reward related to collision, which is set to -1 if the ego vehicle collides and 0 otherwise. $v_{\textrm{lon}}$ is the longitudinal speed of the ego vehicle. $r_{\textrm{fast}}$ is the reward related to running too fast, which is set to -1 if it exceeds the desired speed (8\,$m/s$ here) and 0 otherwise. $r_{\textrm{our}}$ is set to -1 if the ego vehicle runs out of lane, and 0 otherwise. $\alpha$ is the steering angle of ego vehicle in rad. $r_{\textrm{lat}}$ is the reward related to lateral acceleration, which is calculated by $r_{\textrm{lat}}=-|\alpha|v_{\textrm{lon}}^2$. The last constant term is added to prevent the ego vehicle from standing still. 

\subsubsection{Network Architecture}
The parametrized neural networks in our method includes the generative models $p\left(x_t|z_t\right)$ and $p\left(m_t|z_t\right)$, the latent dynamics $p\left(z_{t+1}|z_t,a_t\right)$, the filtering model $q\left(z_{t+1}|z_t,x_{t+1},a_t\right)$ and $q\left(z_1|x_1\right)$, the Q network $Q\left(z_t,a_t\right)$, and the policy network $\pi\left(a_t|z_t\right)$. Here we followed the two-layer hierarchical latent space structure as in \cite{lee2019stochastic}, such that $z_t^1 \in \mathbb{R}^{32}$ and $z_t^2 \in \mathbb{R}^{256}$. Each sensor input size and mask size is $64 \times 64 \times 3$, such that $x_t, m_t \in \left[0,255\right]^{64\times64\times3}$.

$p\left(x_t|z_t\right)$ and $p\left(m_t|z_t\right)$ both consist of 5 deconvolutional layers ((256, 4, 1), (128, 3, 2), (64, 3, 2), (32, 3, 2), and (3, 5, 2), with each tuple means (filters, kernel size, strides)). $p\left(z_{t+1}|z_t,a_t\right)$ consists of two fully connected layers with hidden units number 256, followed by a Gaussian output layer. $q\left(z_{t+1}|z_t,x_{t+1},a_t\right)$ and $q\left(z_1|x_1\right)$ both consist of 5 convolutional layers ((32, 5, 2), (64, 3, 2), (128, 3, 2), (256, 3, 2), and (256, 4, 1), with each tuple means (filters, kernel size, strides)) to first encode the sensor inputs $x_t$ into features of size 256. Then two fully connected layers with hidden units number 256 are followed, with a Gaussian output layer. $Q\left(z_t,a_t\right)$ consists of two fully connected layers with hidden units number 256, followed by a linear output layer. $\pi\left(a_t|z_t\right)$ consists of two fully connected layers with hidden units number 256, followed by a Gaussian layer, and a  tanh bijector.

\subsubsection{Training Details}
At each new episode, the ego vehicle is placed in a random feasible start position in the virtual town. Other vehicles are also located to new random positions. The maximum episode length is 500, the time interval for adjacent frames is 0.1 second. We use a frame skip of 4 for temporal extension, which means the action is fixed for every 4 environment steps. 

The hyperparameters are the same with \cite{haarnoja2018soft}. One gradient step is applied per each skipped frame environment step (e.g, in our case it is one gradient step per every 4 environment steps). The Q network and policy are trained with batch size 256 and learning rate 0.0003. The sequential latent model is trained with batch size 32 and learning rate 0.0001. The length of sequential model used for training is $\tau = 10$. The discount factor $\gamma=0.99$.

\section{Evaluation Results}
During evaluation, we use the same stochastic policy that is used during training. 10 episodes are performed at each evaluation step and the average return is calculated. Same with the training phase, all vehicles are randomly relocated in the whole map for each new episode. No frame skip is performed at the evaluation phase. 

\vspace{-2mm}
\subsection{Variants of Proposed Method}
Besides our proposed method, we also trained and evaluated other two variants of the method, and then compare the three methods:
\subsubsection{\bf{Sensor Inputs and Mask (Proposed)}}
This is our proposed, which takes the sensor inputs and generate the mask.

\subsubsection{\bf{Sensor Inputs Only}}
Here we consider the case that no mask is provided. So only the camera and lidar sensor inputs are inputted and reconstructed. The model learning part is then trained in an unsupervised way without mask labels. 

\subsubsection{\bf{Mask Input Only}}
Assume we already have a good perception and localization system that can accurately detect vehicles, localize ego vehicle, and provide accurate road condition information, we can then directly generate the mask and use it as our input. In this case, only the mask is inputted and reconstructed. Note this method can be regarded as an extension of our paper \cite{chen2019model}, which uses offline data to train a non-sequential variational auto-encoder (VAE) to learn the latent state, and then apply SAC on the latent space. 
 
\vspace{-2mm}


\subsection{Baseline RL Algorithms}
We compare our proposed methods with the following state-of-the-art model-free RL algorithms: 
\subsubsection{\bf{DQN\cite{mnih2015human}}}
DQN is the first deep reinforcement learning method proposed by DeepMind. It uses deep neural network approximate the Q value and uses deep learning to approximate the bellman operation.

\subsubsection{\bf{DDPG\cite{lillicrap2015continuous}}}
DDPG is an actor-critic algorithm on the deterministic policy gradient which is able to handle continuous action spaces. Besides a deep Q network that is approximated with bellman operation, there is a policy network which is optimized with policy gradient.

\subsubsection{\bf{TD3\cite{fujimoto2018addressing}}}
For value-based and actor-critic based RL methods such as DQN and DDPG, function approximation errors will lead to overestimated value estimates and sub-optimal policies. TD3 improves the function approximation errors by taking the minimum value between a pair of critics and delaying policy updates.

\subsubsection{\bf{SAC\cite{haarnoja2018soft}}}
SAC is a fundamentally different RL algorithm compared to the above methods, which is within the MaxEnt RL framework. We have briefly introduced the algorithm in Section~\ref{sec:policylearning}.

\vspace{1mm}

To make a fair comparison, we use the same encoding networks with our proposed method for those baseline algorithms, but now without decoders. We use recurrent neural networks (RNN), since our proposed method also considers time sequence. The type of RNN we use is long short term memory (LSTM), with LSTM size of 40 and output size of 100. 

\vspace{-2mm}
\subsection{Evaluation Results}\label{sec:eval}

\begin{figure}
  \centering
  \includegraphics[width = .48\textwidth]{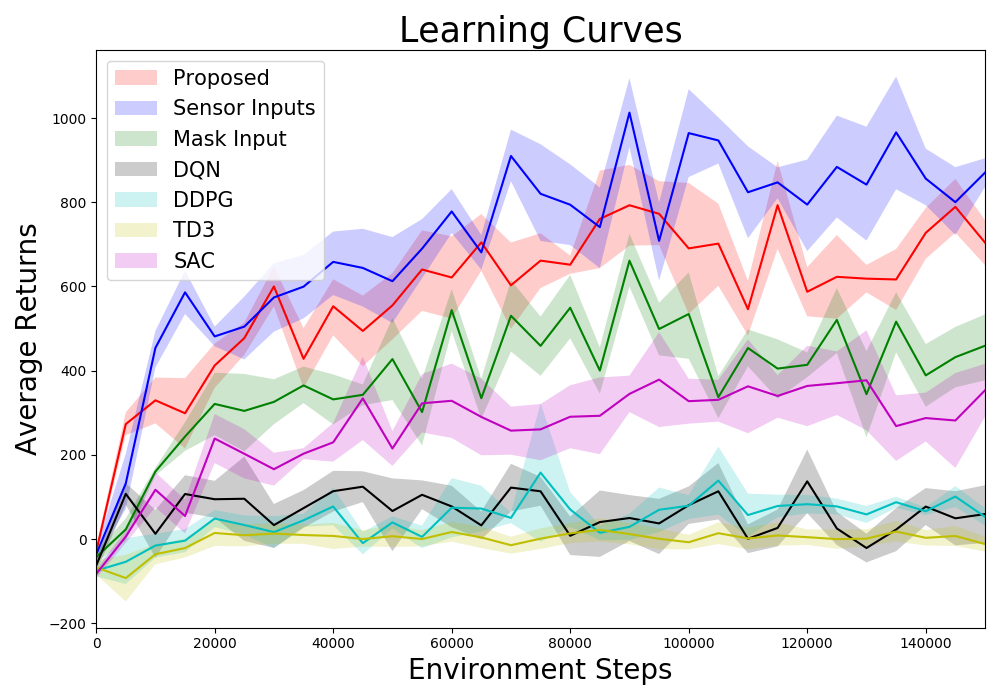}
  \caption{\label{Fig:learning_curve}Comparison of learning curves with baseline RL algorithms. Average returns calculated with 5 trials, each with 10 episodes. Shaded area indicates standard deviation.}
\end{figure}

The performance comparison is shown in Fig.\ref{Fig:learning_curve}. We draw the learning curves composed of average returns (the average discounted cumulative rewards of multiple testing episodes) vs environment steps. We can see that all variants of our proposed method are significantly better than the baselines. Actually, the baselines almost do not work at all. Note that our baselines implemented here are already better than existing RL methods for autonomous driving, which mostly only take front-view camera image, do not consider time sequence, and do not use some state-of-the-art RL algorithms such as SAC and TD3.

\section{Interpretability}
Besides the performance, our proposed method also has significant advantage in terms of intepretability by decoding a semantic mask from the latent state. However, since the baseline RL algorithms do not have a latent space, they are not able to provide an interpretable semantic mask. In this section, we will explain how our method is able to interpret how the autonomous car understand the environment. 

\subsection{Detection \& Localization Functionality}
It is essential to localize the autonomous car and understand the road conditions around the car. Traditionally, this is enabled by a separate localization \& mapping system, which requires the collection of an HD map and designing of complex SLAM~\cite{yurtsever2019survey} algorithms. However, our method is able to obtain all those information within the end-to-end RL training process, without storing any HD maps or manually designing any localization algorithms. 

On the other hand, Object detection is of fundamental importance, as failing to detect road participants and obstacles might lead to serious incidents. The environment model obtained in our method also has the ability to detect surrounding vehicles by fusing camera and lidar sensor inputs.

\begin{figure*}
  \centering
  \includegraphics[width = 1.0\textwidth]{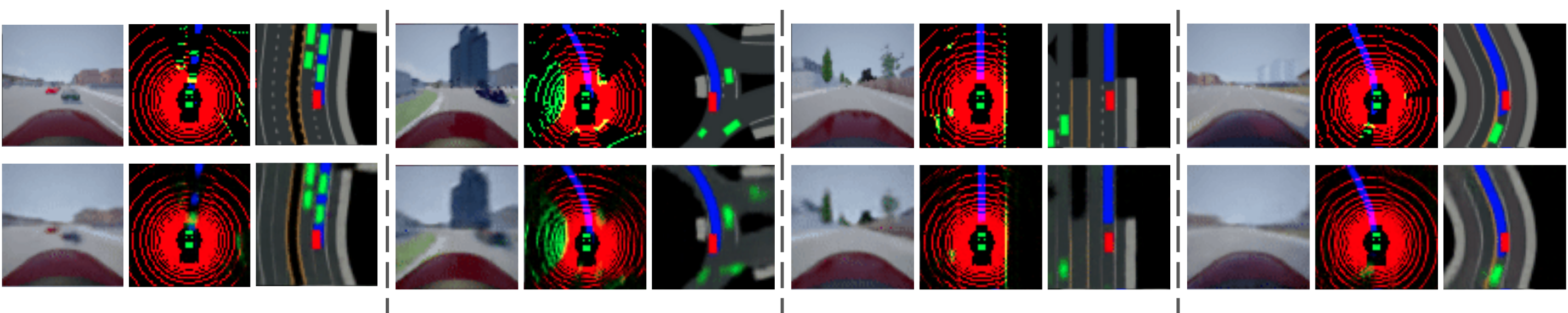}
  \caption{\label{Fig:detect_loc}Randomly sampled frames to illustrate the interpretability of our method. For each sample, left to right: camera, lidar, bird-view image. First row: original sensor inputs and ground truth mask. Second row: reconstructed images. Only the raw camera and lidar images are observed.}
\end{figure*}

Fig.\ref{Fig:detect_loc} shows random sampled frames of the sensor inputs, ground truth masks, and reconstructions during running with the learned model and policy. For each sample, the first row contains the raw sensor inputs and ground truth mask (left to right: camera, lidar, bird-view mask). The second row contains the corresponding reconstructed images from the latent state. Note here only the raw sensor inputs are observed, the ground truth bird-view image is displayed only for comparison. From the reconstructed bird-view mask, we can see that it can accurately locate the ego car and decode the map information (e.g, drivable areas and road markings), even though there is no direct information from the raw sensor inputs indicating the ego car is in an intersection. We can also see that our model can accurately detect the surrounding vehicles (green boxes) given raw camera and lidar observations.

\subsection{Quantified Evaluation}
We also quantify the interpretability of our method by calculating the average pixel difference between the decoded masks and the ground truth masks during massive simulation tests in the virtual city. The metric is defined as:

\begin{equation}
    e=\frac{1}{N}\sum_{i=1}^{N}\frac{\texttt{sum}\left(|\hat{m_i}-m_i|\right)}{W\times H\times C}
\end{equation}
where $\hat{m_i}$ is the predicted mask, $m_i$ is the ground truth mask, $N$ is the number of samples we evaluate. $W$, $H$ and $C$ are the size of the mask image. In our case, $W=H=64$, $C=3$. Values in $m_i$ and $\hat{m_i}$ are RGB values scaled to $\left[0,1\right]$. After evaluating $N=10^4$ frames in the simulation, we got the average pixel difference to be $e=0.032$, which indicates high accuracy when decoding the birdeye semantic mask images.

\subsection{Failure Cases Interpretation}
\begin{figure}
  \centering
  \includegraphics[width = 0.48\textwidth]{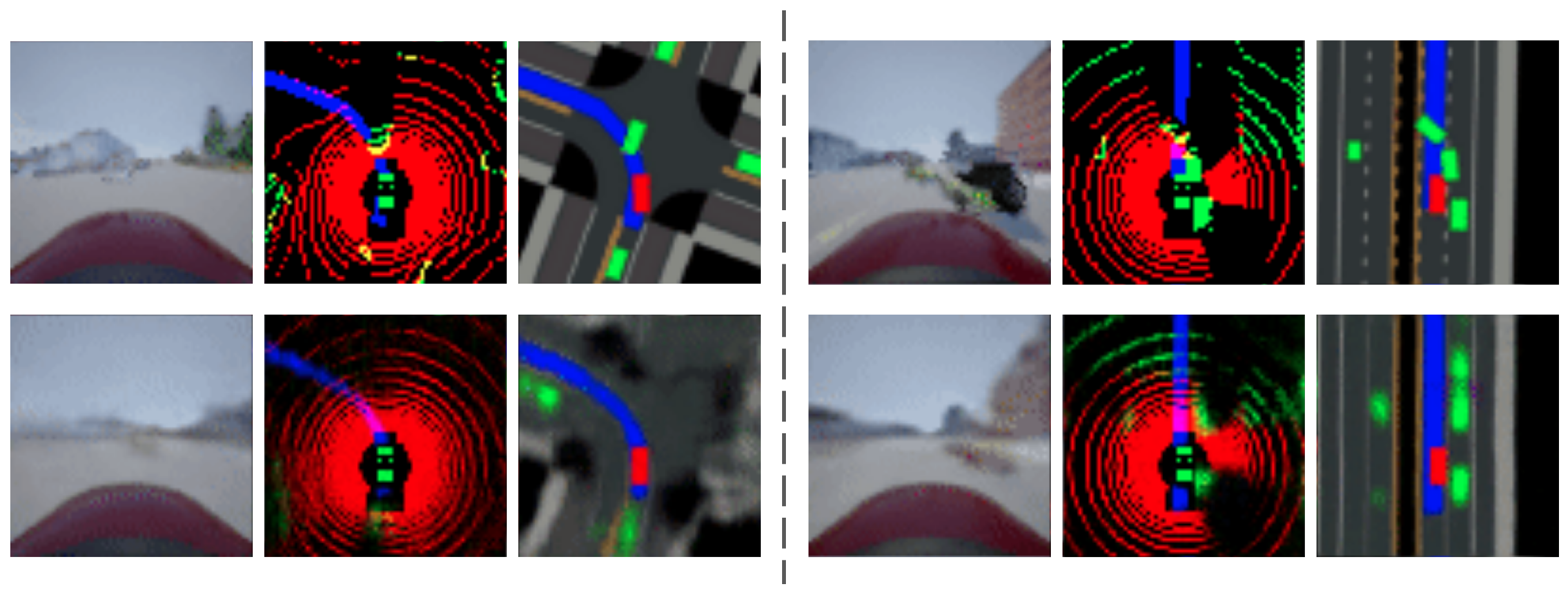}
  \caption{\label{Fig:failure}Examples of failure cases interpretation. }
\end{figure}

Although we can learn significantly better driving policy than baseline RL methods as shown in Section~\ref{sec:eval}, we can still observe some failure cases such as collisions with surrounding vehicles during testing. Our method can help interpret why the agent fails. Fig.\ref{Fig:failure} shows examples of our failure cases interpretation. Same as before, the first row shows the sensor inputs and ground truth masks, while the second row shows the reconstructed sensor inputs and masks. The left example shows a case where the agent collides with another vehicle in an intersection. From the reconstructed mask we can see the agent does not recognize the surrounding vehicle. This might be caused by the low resolution of the sensor inputs, as we can hardly see the vehicle in the camera image. The right example shows a case where the agent collides with a vehicle occupying part of its lane. From the reconstructed mask we can see that although the agent recognize the vehicle, it mistakenly localize it in its own lane. This might because this is a very rare situation and almost all training data is composed of vehicles running in their own lanes.

\section{Conclusions}
In this paper, we proposed an interpretable end-to-end reinforcement learnig algorithm for autonomous driving in urban driving scenarios. The driving policy was learned jointly with a sequential environment model using latent state space. The learned driving policy took camera and lidar images as input, and generated control commands to navigate the autonomous car through urban driving scenarios. The learned environment model provided an interpretable explanation of how the autonomous car understands the driving situation by generating a bird-view semantic mask. The mask was enforced to connect with a certain sub-module in traditional autonomous driving framework, thus providing an explanation of how the learned policy behave in response to current-time environment. The method was implemented and evaluated in CARLA simulator, which was shown to have significantly better performance over classic RL baselines. 

Although our framework is able to provide interpretable explanations about how the model understand the environment, it does not provide any intuition about how it makes the decisions, because the driving policy is obtained in a model-free way. In the future, model-based method will be investigated within in this framework to further improve the performance and interpretability.

\bibliography{refs}

%
\vspace{-3mm}
\begin{IEEEbiography}
[{\includegraphics[width=1in,clip,keepaspectratio]{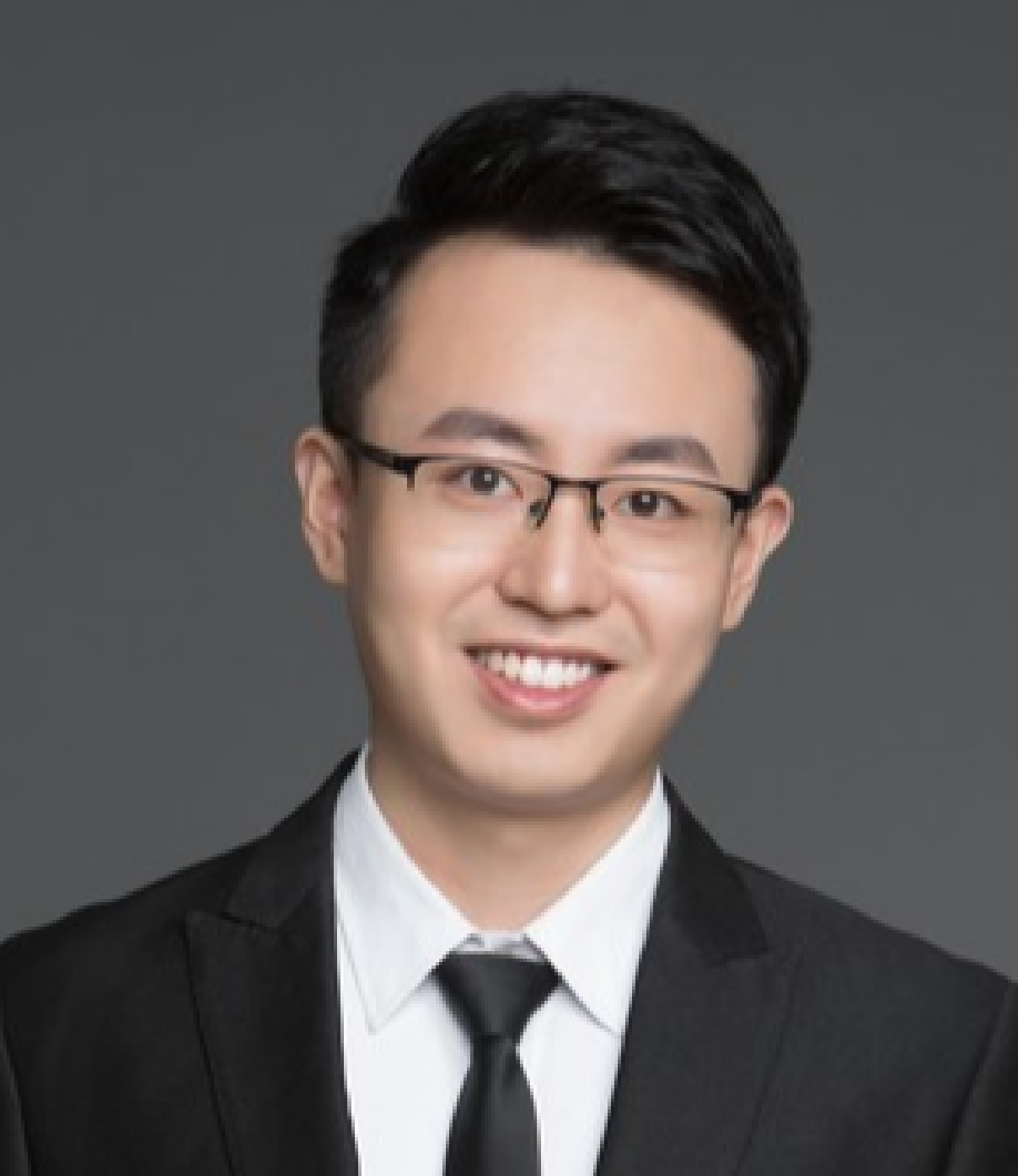}}]{Jianyu Chen} received the B.E. degree from Tsinghua University, China, in 2015. He received the Ph.D. degree working with Prof. Masayoshi Tomizuka at the University of California, Berkeley in 2020. He is working at an intersection of robotics and machine learning to build structured learning systems for intelligent robots which can learn safe, interpretable and scalable perceptual-control policies. Applications of his work mainly focus on autonomous driving. His research interests include motion planning, optimal control, reinforcement learning, imitation learning, and representation learning.
\end{IEEEbiography}
\vspace{-3mm}
\begin{IEEEbiography}
[{\includegraphics[width=1in,clip,keepaspectratio]{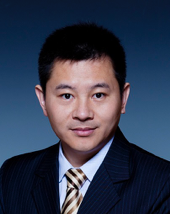}}]{Shengbo Eben Li} received the M.S. and Ph.D. degrees from Tsinghua University in 2006 and 2009. Before joining Tsinghua University, he has worked at Stanford University, University of Michigan, and UC Berkeley. He is now leading Intelligent Driving Lab (iDLab) at Tsinghua University. His active research interests include intelligent vehicles and driver assistance, reinforcement learning and optimal control, distributed control and estimation, etc. He is the author of over 100 peer-reviewed journal/conference papers, and the co-inventor of over 30 patents. Dr. Li was the recipient of Best Paper Award in 2014 IEEE ITS, Best Paper Award in 14th Asian ITS, National Award for Technological Invention of China (2013), Excellent Young Scholar of NSF China (2016), Young Professorship of Changjiang Scholar Program (2016), Tsinghua University Excellent Professorship Award (2017), National Award for Progress in Science and Technology of China (2018), Distinguished Young Scholar of Beijing NSF (2018), etc. He also serves as Board of Governor of IEEE ITS Society, AEs of IEEE ITSM, IEEE Trans ITS, etc.
\end{IEEEbiography}
\vspace{-3mm}

\begin{IEEEbiography}
[{\includegraphics[width=1in,height=1.25in,clip,keepaspectratio]{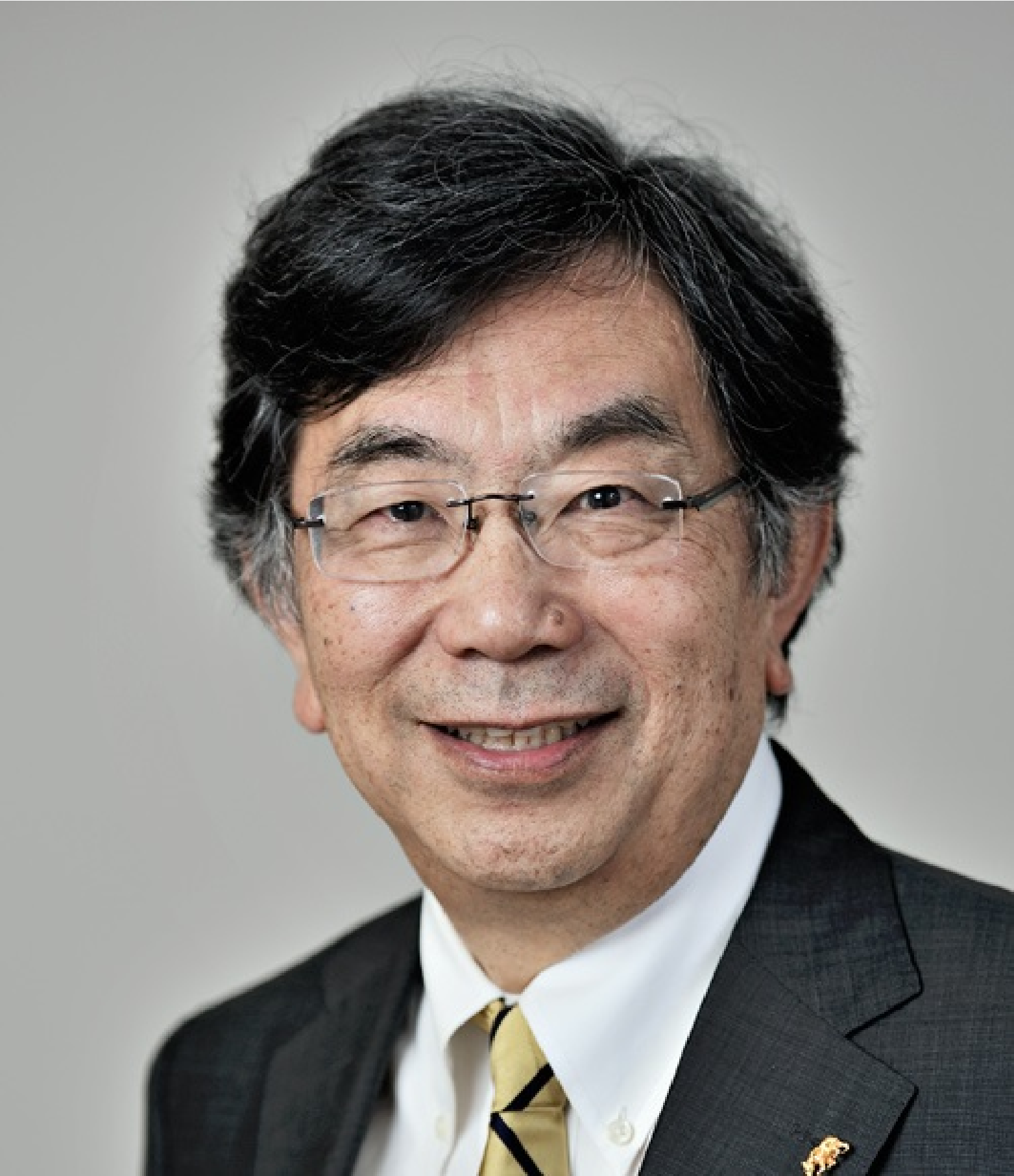}}]{Masayoshi Tomizuka} (M'86-SM'95-F'97-LF'17) received his Ph. D. degree in Mechanical Engineering from MIT in February 1974. In 1974, he joined the faculty of the Department of Mechanical Engineering at the University of California at Berkeley, where he currently holds the Cheryl and John Neerhout, Jr., Distinguished Professorship Chair. His current research interests are optimal and adaptive control, digital control, signal processing, motion control, and control problems related to robotics, precision motion control and vehicles. He served as Program Director of the Dynamic Systems and Control Program of the Civil and Mechanical Systems Division of NSF (2002- 2004). He served as Technical Editor of the ASME Journal of Dynamic Systems, Measurement and Control, J-DSMC (1988-93), and Editor-in-Chief of the IEEE/ASME Transactions on Mechatronics (1997-99). Prof. Tomizuka is a Fellow of the ASME, IEEE and IFAC. He is the recipient of the Charles Russ Richards Memorial Award (ASME, 1997), the Rufus Oldenburger Medal (ASME, 2002) and the John R. Ragazzini Award (2006).  
\end{IEEEbiography}




\end{document}